\documentclass[11pt,a4paper]{article}
\usepackage[hyperref]{emnlp-ijcnlp-2019}
\usepackage{times,subcaption}
\usepackage{latexsym}
\usepackage{graphicx}
\usepackage{url}
\usepackage{booktabs}
\usepackage{amsmath,amsthm,amssymb,pstricks,pst-node}
\usepackage{xspace}

\newsavebox\FrameBox
\newenvironment{Frame}{%
   \par\setbox\FrameBox\hbox\bgroup\minipage{0.45\textwidth}\parskip\baselineskip\ignorespaces
}{%
   \endminipage\egroup\fbox{\box\FrameBox}\par
}
\aclfinalcopy % Uncomment this line for the final submission
\definecolor{dkgreen}{RGB}{0,130,0}

\newcommand{\disceval}{DiscoEval\xspace}
\newcommand{\senteval}{SentEval\xspace}
\newcommand{\elmo}{ELMo\xspace}
\newcommand{\skipthought}{Skip-thought\xspace}
\newcommand{\infersent}{InferSent\xspace}
\newcommand{\dissent}{DisSent\xspace}
\newcommand{\bert}{BERT\xspace}
\newcommand{\bertbase}{BERT-Base\xspace}
\newcommand{\bertlarge}{BERT-Large\xspace}

\newcommand{\spos}{Sentence Position\xspace}
\newcommand{\bso}{Binary Sentence Ordering\xspace}
\newcommand{\dc}{Discourse Coherence\xspace}
\newcommand{\ssp}{Sentence Section Prediction\xspace}

\newcommand{\sposshort}{SP\xspace}

\newcommand{\dcshort}{DC\xspace}

\newcommand{\mingda}[1]{\textcolor{red}{$_{M}$[#1]}}
\newcommand{\todo}[1]{\textcolor{blue}{[TODO: #1]} }
\newcommand{\kevin}[1]{\textcolor{dkgreen}{$_{KG}$[#1]}}
\newcommand{\zewei}[1]{\textcolor{red}{$_{Z}$[#1]}}

\renewcommand{\mingda}[1]{}
\renewcommand{\todo}[1]{}
\renewcommand{\kevin}[1]{}
\renewcommand{\zewei}[1]{}

\newenvironment{itemizesquish}{\begin{list}{\labelitemi}{\setlength{\itemsep}{-0.2em}\setlength{\labelwidth}{0.5em}\setlength{\leftmargin}{\labelwidth}
\addtolength{\leftmargin}{\labelsep}}}{\end{list}}

\title{Evaluation Benchmarks and Learning Criteria\\for Discourse-Aware Sentence Representations}

\author{Mingda Chen$^{2}$\thanks{~~Equal contribution. Listed in alphabetical order.} \quad Zewei Chu$^{1*}$\quad Kevin Gimpel$^{2}$ \\
$^{1}$University of Chicago, IL, USA\\
$^{2}$Toyota Technological Institute at Chicago, IL, USA\\
  \texttt{\{mchen,kgimpel\}@ttic.edu,zeweichu@uchicago.edu}
}

\begin{document}
\maketitle
\begin{abstract}

Prior work on pretrained sentence embeddings and benchmarks focuses on the capabilities of representations for stand-alone sentences. 
We propose \disceval, a test suite of tasks to evaluate whether sentence representations include information about the role of a sentence in its discourse context. We also propose a variety of training objectives that make use of natural annotations from Wikipedia to build sentence encoders capable of modeling discourse information. 
We benchmark sentence encoders trained with our proposed objectives, as well as other popular pretrained sentence encoders, on \disceval and other sentence evaluation tasks.
Empirically, we show that these training objectives help to encode different aspects of information from the surrounding document structure. Moreover, \bert~\cite{devlin2018bert} and \elmo~\cite{peters2018elmo} demonstrate strong performance across \disceval tasks with individual hidden layers showing different characteristics.\footnote{Data processing and evaluation scripts are available at \href{https://github.com/ZeweiChu/DiscoEval}{\nolinkurl{https://github.com/ZeweiChu/DiscoEval}}.}

\end{abstract}

\section{Introduction}

Pretrained sentence representations have been found useful in various downstream tasks such as visual question answering~\citep{tapaswi2016movieqa}, script inference~\citep{pichotta-mooney-2016-using}, and information retrieval~\citep{le2014distributed, Palangi:2016:DSE:2992449.2992457}. Benchmark datasets~\citep{adi2017fine,conneau2018senteval,wang2018glue,wang2019superglue} have been proposed to evaluate the encoded knowledge, where the focus has been primarily on natural language understanding capabilities of the  representation of a stand-alone sentence, such as its semantic roles, rather than the broader context in which it is situated.

\begin{figure}[t]
\small
\centering
\includegraphics[scale=0.5]{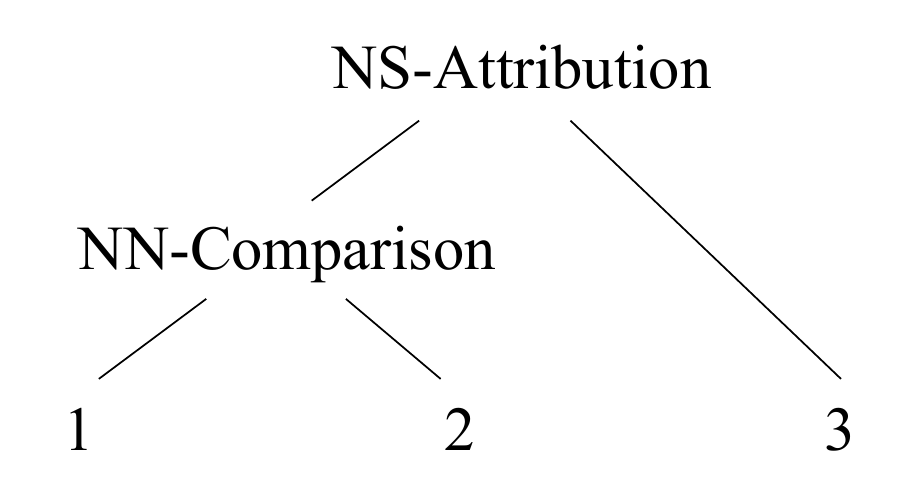}
\vspace{-0.1in}
\captionsetup{font=small}
\caption*{[The European Community's consumer price index rose a provisional 0.6\% in September from August]\textsubscript{1}
[and was up 5.3\% from September 1988,]\textsubscript{2}
[according to Eurostat, the EC's statistical agency.]\textsubscript{3}}
\vspace{-0.07in}
\captionsetup{font=10pt}
\caption{An RST discourse tree from the RST Discourse Treebank. ``N'' represents ``nucleus'', containing basic information for the relation. ``S'' represents ``satellite'', containing additional information about the nucleus.}\label{rst-example}
\vspace{-0.2in}
\end{figure}

In this paper, we seek to incorporate and evaluate discourse knowledge in general purpose sentence representations. 
A discourse is a coherent, structured group of sentences that acts as a fundamental type of structure in natural language~\cite{Jurafsky:2009:SLP:1214993}. A discourse structure is often characterized %exhibited 
by the arrangement of semantic elements across multiple sentences, such as entities and pronouns. The simplest such arrangement (i.e., linearly-structured) can be understood as sentence ordering, where the structure is manifested in the timing of introducing entities. Deeper discourse structures use more complex relations among sentences (e.g., tree-structured; see Figure~\ref{rst-example}).

Theoretically, discourse structures have been approached through Centering Theory~\cite{grosz1995centering} for studying distributions of entities across text and Rhetorical Structure Theory (RST; \citealp{mann1988rhetorical}) for modelling the logical structure of natural language via discourse trees. Researchers have found modelling discourse useful in a range of tasks~\cite{guzman-etal-2014-using,narasimhan-barzilay-2015-machine,liu-lapata-2018-learning,pan-etal-2018-discourse}, including summarization~\cite{gerani-etal-2014-abstractive}, text classification~\cite{ji-smith-2017-neural}, and text generation~\cite{bosselut-etal-2018-discourse}.

In this paper, we propose \disceval, a task suite designed to evaluate discourse-related knowledge in pretrained sentence representations. 
\disceval comprises 7 task groups covering multiple domains, including Wikipedia, stories, dialogues, and scientific literature. 
The tasks are probing tasks~\cite[\emph{inter alia}]{shi-etal-2016-string,adi2017fine,belinkov-etal-2017-evaluating,peters-etal-2018-dissecting,conneau-etal-2018-cram,poliak-etal-2018-collecting,tenney2018what,liu-etal-2019-linguistic,ettinger2019bert,mchen-enteval-19} based on sentence ordering, annotated discourse relations, and discourse coherence. The data is either generated semi-automatically or based on human annotations~\citep{carlson-etal-2001-building,Prasad08pdtb,lin09pdtb,kummerfeld-etal-2019-large}.

We also propose a set of novel multi-task learning objectives building upon standard pretrained sentence encoders, which rely on the assumption of distributional semantics of text.
These objectives depend only on the natural structure in structured document collections like Wikipedia.

Empirically, we benchmark our models and several popular sentence encoders on \disceval and \senteval~\citep{conneau2018senteval}. We find that our proposed training objectives help the models capture different characteristics in the sentence representations.
Additionally, we find that \elmo shows strong performance on \senteval, whereas \bert performs the best among the pretrained embeddings on \disceval. Both \bert and \skipthought vectors~\citep{Kiros2015skipthought}, which have training losses explicitly related to surrounding sentences, perform much stronger compared to their respective prior work, demonstrating the effectiveness of incorporating losses that make use of broader context. Through per-layer analysis, we also find that for both \bert and \elmo, deep layers consistently outperform shallower ones on \disceval, showing different trends from \senteval where the shallow layers have the best performance. 

\section{Related Work}
 
Discourse modelling and discourse parsing have a rich history~\cite[\emph{inter alia}]{marcu2000discourse,barzilay2008modeling,zhou10implicit,kalchbrenner13discoursecompositionality,ji2015one,li2017opendiscourse,wang-etal-2018-toward,liu-etal-2018-discourse,lin-etal-2019-unified}, much of it based on recovering linguistic annotations of discourse structure.

Several researchers have defined tasks related to discourse structure, including sentence ordering~\cite{chen2016neural,logeswaran2016sentence,cui2018deep}, sentence clustering~\cite{wang2018storysalad}, and disentangling textual threads~\citep{elsner-charniak-2008-talking,elsner-charniak-2010-disentangling,lowe-etal-2015-ubuntu,mehri-carenini-2017-chat,N18-1164,kummerfeld-etal-2019-large}.

There is a great deal of prior work on pretrained representations~\cite[\emph{inter alia}]{le2014distributed,Kiros2015skipthought,hill2016fastsent,wieting-16-full,NIPS2017_7209,gan-etal-2017-learning,peters2018elmo,logeswaran2018an,devlin2018bert,tang-de-sa-2019-exploiting,yang2019xlnet,liu2019roberta}.~\skipthought vectors form an effective architecture for general-purpose sentence embeddings. The model encodes a sentence to a vector representation, and then predicts the previous and next sentences in the discourse context. Since Skip-thought performs well in downstream evaluation tasks, we use this neighboring-sentence objective as a starting point for our models.

There is also work on incorporating discourse related objectives into the training of sentence representations. \citet{jernite2017discourse} propose binary sentence ordering, conjunction prediction (requiring manually-defined conjunction groups), and next sentence prediction. Similarly, \citet{sileo-etal-2019-mining} and \citet{nie-etal-2019-dissent} create training datasets automatically based on discourse relations provided in the Penn Discourse Treebank~(PDTB; \citealp{lin09pdtb}). 

Our work differs from prior work in that we propose a general-purpose pretrained sentence embedding evaluation suite that covers multiple aspects of discourse knowledge and we propose novel training signals based on document structure, including sentence position and section titles, without requiring additional human annotation. 

\section{Discourse Evaluation}

We propose \disceval, a test suite of 7 tasks to evaluate whether sentence representations include semantic information relevant to discourse processing. 
Below we describe the tasks and datasets, as well as the evaluation framework. 
We closely follow the SentEval sentence embedding evaluation suite, in particular its supervised sentence and sentence pair classification tasks, which use predefined neural architectures with slots for fixed-dimensional sentence embeddings.  
All \disceval tasks are modelled by logistic regression unless otherwise stated in later sections. 

We also experimented with adding hidden layers to the \disceval classification models. However, we find simpler linear classifiers to provide a clearer comparison among sentence embedding methods. More complex classification models lead to noisier results, as more of the modelling burden is shifted to the optimization of the classifiers. Hence we decide to evaluate the sentence embeddings with simple classification models. 

In the rest of this section, we will use $[\cdot,\cdot,\cdots]$ to denote concatenation of vectors, $\odot$ for element-wise multiplication, and $\vert\cdot\vert$ for element-wise absolute value.

\subsection{Discourse Relations}

As the most direct way to probe discourse knowledge, we consider the task of predicting annotated discourse relations among sentences. We use two human-annotated datasets: the RST Discourse Treebank~(RST-DT; \citealp{carlson-etal-2001-building}) and the Penn Discourse Treebank~(PDTB; \citealp{Prasad08pdtb}). They have different labeling schemes. PDTB provides discourse markers for adjacent sentences, whereas RST-DT offers document-level discourse trees, which recently was used to evaluate discourse knowledge encoded in document-level models~\cite{ferracane-etal-2019-evaluating}. The difference allows us to see if the pretrained representations capture local or global information about discourse structure.

More specifically, as shown in Figure~\ref{rst-example}, in RST-DT, text is segmented into basic units, elementary discourse units (EDUs), upon which a discourse tree is built recursively. Although a relation can take multiple units, we follow prior work~\cite{ji-eisenstein-2014-representation} to use right-branching trees for non-binary relations to binarize the tree structure and use the 18 coarse-grained relations defined by \citet{carlson-etal-2001-building}. 

When evaluating pretrained sentence encoders on RST-DT, we first encode EDUs into vectors, then use averaged vectors of EDUs of subtrees as the representation of the subtrees. The target prediction is the label of nodes in discourse trees and the input to the classifier is $[x_\text{left}, x_\text{right}, x_\text{left} \odot x_\text{right}, |x_\text{left} - x_\text{right}|]$, where $x_\text{left}$ and $x_\text{right}$ are vector representations of the left and right subtrees respectively. For example, the input for target ``NN-Attribution'' in Figure~\ref{rst-example} would be $x_\text{left}=\frac{x_1+x_2}{2}$, $x_\text{right}=x_3$, where $x_i$ is the encoded representation for the $i$th EDU in the text. We use the standard data splits, where there are 347 documents for training and 38 documents for testing. We choose 35 documents from the training set to serve as a validation set.

For PDTB, we use a pair of sentences to predict discourse relations.
Following \citet{lin09pdtb}, we focus on two kinds of relations from PDTB: explicit (PDTB-E) and implicit (PDTB-I). The sentence pairs with explicit relations are two consecutive sentences with a particular connective word in between. 
Figure~\ref{fig:example-pdtb-e} is an example of an explicit relation.  

\begin{figure}[t]
    \small
    \centering
    \begin{Frame}
        1. In any case, the brokerage firms are clearly moving faster to create new ads than they did in the fall of 1987. \\
        2. $[$But$]$ it remains to be seen whether their ads will be any more effective. \\
        label: Comparison.Contrast
    \end{Frame}
    \caption{Example in the PDTB explicit relation task. The words in $[]$ are taken out from input sentence 2. }

    \label{fig:example-pdtb-e}
\end{figure}

In the PDTB, annotators insert an implicit connective between adjacent sentences to reflect their relations, if such an implicit relation exists. 
Figure~\ref{fig:example-pdtb-i} shows an example of an implicit relation.
The PDTB provides a three-level hierarchy of relation tags. In \disceval, we use the second level of types~\citep{lin09pdtb}, as they provide finer semantic distinctions compared to the first level. To ensure there is a reasonable amount of evaluation data, we use sections 2-14 as training set, 15-18 as development set, and 19-23 as test set. In addition, we filter out categories that have less than 10 instances. This leaves us 12 categories for explicit relations and 11 for implicit ones. Category names are listed in the supplementary material.
\begin{figure}
    \small
    \centering
    \begin{Frame}
        1. ``A lot of investor confidence comes from the fact that they can speak to us,'' he says. \\
        2. $[$so$]$ ``To maintain that dialogue is absolutely crucial.'' \\
        label: Contingency.Cause
    \end{Frame}
    \caption{Example in the PDTB implicit relation task. }
    \label{fig:example-pdtb-i}
\end{figure}

We use the sentence embeddings to infer sentence relations with supervised training. As input to the classifier, we encode both sentences to vector representations $x_1$ and $x_2$, concatenated with their element-wise product and absolute difference: $[x_1, x_2, x_1 \odot x_2, |x_1 - x_2|]$.

\subsection{\spos (SP)}

We create a task that we call \spos. It can be seen as way to probe the knowledge of linearly-structured discourse, where the ordering corresponds to the timings of events. When constructing this dataset, we take five consecutive sentences from a corpus, randomly move one of these five sentences to the first position, and ask models to predict the true position of the first sentence in the modified sequence.

We create three versions of this task, one for each of the following three domains: the first five sentences of the introduction section of a Wikipedia article~(Wiki), the ROC Stories corpus~(ROC; \citealp{rocstory}), and the first 5 sentences in the abstracts of arXiv papers~(arXiv; \citealp{chen2016neural}).
Figure~\ref{fig:example-spos} shows an example of this task for the ROC Stories domain. The first sentence should be in the fourth position among these sentences. To make correct predictions, the model needs to be aware of both typical orderings of events as well as how events are described in language. In the example shown, Bonnie's excitement comes from her imagination so it must happen after she picked up the jeans and tried them on but right before she realized the actual size. 

\begin{figure}
    \small
    \centering
    \begin{Frame}
        - She was excited thinking she must have lost weight.  \\
        - Bonnie hated trying on clothes.    \\
        - She picked up a pair of size 12 jeans from the display.    \\
        - When she tried them on they were too big! \\
        - Then she realized they actually size 14s, and 12s.
    \end{Frame}
    \caption{Example from the ROC Stories domain of the \spos task. The first sentence should be in the fourth position.}
    \vspace{-0.1in}
    \label{fig:example-spos}
\end{figure}

To train classifiers for these tasks, we do the following. 
We first encode the five sentences
to vector representations $x_i$. As input to the classifier, we include $x_1$ and the concatenation of $x_1 - x_i$ for all $i$: 
$[x_1, x_1 - x_2, x_1 - x_3, x_1 - x_4, x_1 - x_5]$.

\subsection{\bso (BSO)}

Similar to sentence position prediction, \bso (BSO) is a binary classification task to determine the order of two sentences. The fact that BSO only has a pair of sentences as input makes it different from \spos, where there is more context, and we hope that BSO can evaluate the ability of capturing local discourse coherence in the given sentence representations. The data comes from the same three domains as \spos, and each instance is a pair of consecutive sentences.

Figure~\ref{fig:example-bso} shows an example from the arXiv domain of the \bso task. The order of the sentences in this instance is incorrect, as the ``functions'' are referenced before they are introduced. To detect the incorrect ordering in this example, the encoded representations need to be able to provide information about new and old information in each sentence.

\begin{figure}
    \small
    \centering
    \begin{Frame}
        1. These functions include fast and synchronized response to environmental change, or long-term memory about the transcriptional status.\\
        2. Focusing on the collective behaviors on a population level, we explore potential regulatory functions this model can offer.
    \end{Frame}
    \caption{Example from the arXiv domain of the \bso task (incorrect ordering shown). }
    \vspace{-0.1in}
    \label{fig:example-bso}
\end{figure}

To form the input when training classifiers, we concatenate the embeddings of both sentences with their element-wise difference: $[x_1, x_2, x_1 - x_2]$.

\subsection{\dc (DC)}

Inspired by prior work on chat disentanglement~\cite{elsner-charniak-2008-talking,elsner-charniak-2010-disentangling} and sentence clustering~\citep{wang2018storysalad}, we propose a sentence disentanglement task. The task is %. is a binary classification task 
to determine whether a sequence of six sentences forms a coherent paragraph. We start with a coherent sequence of six sentences, then randomly replace one of the sentences (chosen uniformly among positions 2-5) with a sentence from another discourse. 
This task, which we call \dc (DC), is a binary classification task and the datasets are balanced between positive and negative instances.

We use data from two domains for this task: Wikipedia and the Ubuntu IRC channel.\footnote{\href{https://irclogs.ubuntu.com/}{\nolinkurl{irclogs.ubuntu.com/}}} 
For Wikipedia, we begin by choosing a sequence of six sentences from a Wikipedia article. For purposes of choosing difficult distractor sentences, we use the Wikipedia categories of each document as an indication of its topic.
To create a negative instance, we randomly sample a sentence from another document with a similar set of categories (measured by the percentage of overlapping categories). This sampled sentence replaces one of the six consecutive sentences in the original sequence. When splitting the train, development, and test sets, we ensure there are no overlapping documents among them.

Our proposed dataset differs from the sentence clustering task of \citet{wang2018storysalad} in that it preserves sentence order and does not anonymize or lemmatize words, because they play an important role in conveying information about discourse coherence. 

For the Ubuntu domain, we use the human annotations of conversation thread structure from \citet{kummerfeld-etal-2019-large} to provide us with a coherent sequence of utterances. We filter out sentences by heuristic rules to avoid overly technical and unsolvable cases. The negative sentence is randomly picked from other conversations. Similarly, when splitting the train, development, and test sets, we ensure there are no overlapping conversations among them.

Figure~\ref{fig:example-dc} is an instance of the Wikipedia domain of the \dc task. This instance is not coherent and the boldfaced text is from a different document. The incoherence can be found either by comparing characteristics of the entity being discussed or by the topic of the sentence group. %Evidently, 
Solving this task is non-trivial as it may require the ability to perform inference across multiple sentences.

\begin{figure}
    \small
    \centering
    \begin{Frame}
        1. It is possible he was the youngest of the family as the name ``Sextus'' translates to sixth in English implying he was the sixth of two living and three stillborn brothers.	\\
        2. According to Roman tradition, his rape of Lucretia was the precipitating event in the overthrow of the monarchy and the establishment of the Roman Republic.	\\
        3. Tarquinius Superbus was besieging Ardea, a city of the Rutulians.	\\
        4. The place could not be taken by force, and the Roman army lay encamped beneath the walls.\\
        5. \textbf{He was soon elected to the Academy's membership (although he had to wait until 1903 to be elected to the Society of American Artists), and in 1883 he opened a New York studio, dividing his time for several years between Manhattan and Boston.} \\
        6. As nothing was happening in the field, they mounted their horses to pay a surprise visit to their homes.
    \end{Frame} % it's Frame not frame (capitalization matters)
    \caption{An example from the Wikipedia domain of the \dc task. This sequence is not coherent; the boldface sentence was substituted in for the true fifth sentence from another article. }
    \vspace{-0.1in}
    \label{fig:example-dc}
\end{figure}

In this task, we encode all sentences to vector representations and concatenate all of them ($[x_1, x_2, x_3, x_4, x_5, x_6]$) as input to the classification model. Note that in this task, we use a hidden layer of 2000 dimensions with sigmoid activation in the classification model, as this is necessary for the classifier to use features based on multiple inputs simultaneously given the simple concatenation as input. We could have developed richer ways to encode the  input so that a linear classifier would be feasible (e.g., use the element-wise products of all pairs of sentence embeddings), but we wish to keep the input dimensionality of the classifier small enough that the classifier will be learnable given fixed sentence embeddings and limited training data.

\subsection{\ssp (SSP)}

The \ssp (SSP) task is defined as determining the section of a given sentence. The motivation behind this task is that sentences within certain sections typically exhibit similar patterns because of the way people write coherent text. The pattern can be found based on connectives or specificity of a sentence. For example, ``Empirically'' is usually used in the abstract or introduction sections in scientific writing. 

We construct the dataset from PeerRead~\cite{peerread}, which consists of scientific papers from a variety of fields. 
The goal is to predict whether or not a sentence belongs to the Abstract section. 
After eliminating sentences that are too easy for the task (e.g., equations), we randomly sample sentences from the Abstract or from a section in the middle of a paper.\footnote{We avoid sentences from the Introduction or Conclusion sections to make the task more solvable.}
Figure~\ref{fig:example-ssp} shows two sentences from this task, where the first sentence is more general and from an Abstract whereas the second is more specific and is from another section. In this task, the input to the classifier is simply the sentence embedding.

 \begin{figure}[t]
    \small
    \centering
    \begin{Frame}
        1. The theory behind the SVM and the naive Bayes classifier is explored.\\
        2. This relocation of the active target may be repeated an arbitrary number of times.
    \end{Frame}
    \caption{Examples from \ssp. The first is from an Abstract while the second is not. }
    \label{fig:example-ssp}
\end{figure}

Table~\ref{table:dataset-stats} shows the number of instances in each \disceval task introduced above. 

\begin{table}[t]
\small
\centering
\setlength{\tabcolsep}{4pt}
\begin{tabular}{c|ccccc}
\toprule
Task & PDTB-E & PDTB-I & Ubuntu & RST-DT & Others \\
\midrule
Train & 9383 & 8693 & 5816 & 17051 & 10000 \\
Dev. & 3613 & 2972 & 1834 & 2045 & 4000 \\
Test & 3758 & 3024 & 2418 & 2308 & 4000 \\
\bottomrule
\end{tabular}
\vspace{-0.05cm}
\caption{\label{table:dataset-stats} Size of datasets in \disceval.}
\vspace{-0.1cm}
\end{table}

\section{Models and Learning Criteria}
\label{sec:model}

Having described \disceval, we now discuss methods for incorporating discourse information into sentence embedding training. All models in our experiments are composed of a single encoder and multiple decoders.
The encoder, parameterized by a bidirectional Gated Recurrent Unit~(BiGRU; \citealp{chung2014empirical}), encodes the sentence, either in training or in evaluation of the downstream tasks, to a fixed-length vector representation (i.e., the average of the hidden states across positions).

The decoders take the aforementioned encoded sentence representation, and predict the targets we define in the sections below. We first introduce Neighboring Sentence Prediction, the loss for our baseline model. We then propose additional training losses to encourage our sentence embeddings to capture other context information. 

\subsection{Neighboring Sentence Prediction (NSP)}

Similar to prior work on sentence embeddings~\cite{Kiros2015skipthought,hill2016fastsent}, 
we use an encoded sentence representation to predict its surrounding sentences. 
In particular, we predict the immediately preceding and succeeding sentences.
All of our sentence embedding models use this loss. Formally, the loss is defined as

\begin{equation}
    \text{NSP}=-\log p_\theta(s_{t-1}\vert s_t)-\log p_\phi(s_{t+1}\vert s_t)\nonumber 
\end{equation}
\noindent where we parameterize $p_\theta$ and $p_\phi$ as separate feedforward neural networks and compute the log-probability of a target sentence using its bag-of-words representation.

\subsection{Nesting Level (NL)}

A table of contents serves as a high level description of an article, outlining its organizational structure. Wikipedia articles, for example, contain rich tables of contents with many levels of hierarchical structure. The ``nesting level'' of a sentence (i.e., how many levels deep it resides) provides information about its role in the overall discourse. 
To encode this information into our sentence representations, we introduce a discriminative loss to predict a sentence's nesting level in the table of contents:
\begin{equation}
    \text{NL}=-\log p_\theta(l_t\vert s_t)\nonumber
\end{equation}
\noindent where $l_t$ represents the nesting level of the sentence $s_t$ and $p_\theta$ is parameterized by a feedforward neural network. Note that sentences within the same paragraph share the same nesting level. 
In Wikipedia, there are up to 7 nesting levels.

\subsection{Sentence and Paragraph Position (SPP)}
Similar to nesting level, we add a loss based on using the sentence representation to predict its position in the paragraph and in the article. 
The position of the sentence can be a strong indication of 
the relations between the topics of the current sentence and the topics in the entire article. 
For example, the first several sentences often cover the general topics to be discussed more thoroughly in the following sentences. To encourage our sentence embeddings to capture such information, 
we define a position prediction loss
\begin{equation}
    \text{SPP}=-\log p_\theta(sp_t\vert s_t) - \log p_\phi(pp_t\vert s_t)\nonumber
\end{equation}
\noindent where $sp_t$ is the sentence position of $s_t$ within the current paragraph and $pp_t$ is the position of the current paragraph in the whole document.

\subsection{Section and Document Title (SDT)}
Unlike the previous position-based losses, this loss makes use of section and document titles, which gives the model more direct access to the topical information at different positions in the document. The loss is defined as
\begin{equation}
    \text{SDT}=-\log p_\theta(st_t\vert s_t)-\log p_\phi(dt_t\vert s_t)\nonumber
\end{equation}
\noindent Where $st_t$ is the section title of sentence $s_t$, $dt_t$ is the document title of sentence $s_t$, and $p_\theta$ and $p_\phi$ are two different bag-of-words decoders.

\section{Experiments}
\subsection{Setup}

We train our models on Wikipedia as it is a knowledge rich textual resource and has consistent structures over all documents. Details on hyperparameters are in the supplementary material. When evaluating on \disceval,
we encode sentences with pretrained sentence encoders. Following \senteval, we freeze the sentence encoders and only learn the parameters of the downstream classifier. The ``Baseline'' row in Table~\ref{table:results} are embeddings trained with only the NSP loss. The subsequent rows are trained with extra losses defined in Section~\ref{sec:model} in addition to the NSP loss.

\begin{table*}[t]
    \small
    \centering
\begin{tabular}{ccccc|cccccccc}
\toprule
& \multicolumn{4}{c|}{ SentEval } & \multicolumn{8}{c}{ \disceval } \\
& USS & SSS & SC & Probing & SP & BSO & DC & SSP & PDTB-E & PDTB-I & RST-DT & avg. \\
\midrule
\skipthought & 41.7 & 81.2 & 78.4 & 70.1 & 47.5 & 64.6 & 55.2 & 77.5 & 39.3 & 40.2 & \textbf{59.7} & 54.8 \\
\infersent & \textbf{63.4} & \textbf{83.3} & 79.7 & 71.8 & 45.8 & 62.9 & 56.3 & 62.2 & 37.3 & 38.8 & 52.3 & 50.8 \\
\dissent & 50.0 & 79.2 & 80.5 & 74.0 & 47.7 & 64.9 & 54.8 & 62.2 & 42.2 & 40.7 & 57.8 & 52.9 \\
\elmo & 60.9 & 77.6 & 80.8 & 74.7 & 47.8 & 65.6 & \textbf{60.7} & 79.0 & 41.3 & 41.8 & 57.5 & 56.2 \\
BERT-Base & 30.1 & 66.3 & 81.4 & 73.9 & 53.1 & 68.5 & 58.9 & 80.3 & 41.9 & 42.4 & 58.8 & 57.7 \\
BERT-Large & 43.6 & 70.7 & \textbf{83.4} & \textbf{75.0} & \textbf{53.8} & \textbf{69.3} & 59.6 & \textbf{80.4} & \textbf{44.3} & \textbf{43.6} & 59.1 & \textbf{58.6} \\
\midrule
Baseline (NSP) & 57.8 & 77.1 & 77.0 & 70.6 & 47.3 & 63.8 & \underline{61.0} & 77.8 & 36.5 & 39.1 & \underline{56.7} & 54.6 \\
+ SDT & \underline{59.0} & 77.3 & 76.8 & 69.7 & 45.8 & 62.9 & 60.3 & 78.0 & 36.6 & 39.1 & 55.7 & 54.1 \\
+ SPP & 56.0 & 77.5 & \underline{77.4} & \underline{70.7} & 48.4 & \underline{65.3} & 60.2 & 78.4 & \underline{38.1} & 39.9 & 56.4 & 55.2 \\
+ NL & 56.7 & \underline{78.2} & 77.2 & 70.6 & 46.9 & 64.0 & \underline{61.0} & \underline{78.9} & 37.6 & 39.9 & 56.5 & 55.0 \\
+ SPP + NL & 55.4 & 76.7 & 77.0 & 70.4 & \underline{48.5} & 64.7 & 59.9 & \underline{78.9} & 37.8 & \underline{40.5} & \underline{56.7} & \underline{55.3} \\
+ SDT + NL & 58.5 & 76.9 & 77.2 & 70.2 & 46.1 & 63.0 & 60.8 & 78.1 & 36.7 & 38.1 & 56.2 & 54.1 \\
+ SDT +SPP & 58.4 & 77.4 & 76.6 & 70.2 & 46.5 & 63.9 & 60.4 & 77.6 & 35.2 & 38.6 & 56.3 & 54.1 \\
ALL & 58.8 & 76.3 & 77.0 & 70.2 & 46.1 & 63.7 & 60.0 & 78.6 & 36.3 & 37.6 & 55.3 & 53.9 \\
\bottomrule
\end{tabular}
\caption{Results for \senteval and \disceval. The highest number in each column is boldfaced. The highest number for our models in each column is underlined. ``All'' uses all four losses. ``avg.'' is the averaged accuracy for all tasks in \disceval. 
}
  \label{table:results}
  \vspace{-0.1in}
\end{table*}%

Additionally, we benchmark several popular pretrained sentence encoders on \disceval, including~\skipthought,\footnote{\href{https://github.com/ryankiros/skip-thoughts}{\nolinkurl{github.com/ryankiros/skip-thoughts}}}~\infersent~\cite{infersent},\footnote{\href{https://github.com/facebookresearch/InferSent}{\nolinkurl{github.com/facebookresearch/InferSent}}}~\dissent~\cite{nie-etal-2019-dissent},\footnote{\href{https://github.com/windweller/DisExtract}{\nolinkurl{github.com/windweller/DisExtract}}}~\elmo,\footnote{\href{https://github.com/allenai/allennlp}{\nolinkurl{github.com/allenai/allennlp}}} and~\bert.\footnote{\href{https://github.com/huggingface/pytorch-pretrained-BERT}{\nolinkurl{github.com/huggingface/pytorch-pretrained-BERT}}} For \elmo, we use the averaged vector of all three layers and time steps as the sentence representations. For \bert, we use the averaged vector at the position of the ``[CLS]'' token across all layers. We also evaluate per-layer performance for both models in Section~\ref{sec:bert-elmo-per-layer-perf}.

When reporting results for \senteval, we compute the averaged Pearson correlations for Semantic Textual Similarity tasks from 2012 to 2016~\cite{agirre2012semeval,diab2013eneko,agirre2014semeval,agirre2015semeval,agirre2016semeval}. We refer to the average as unsupervised semantic similarity (USS) since those tasks do not require training data. 
We compute the averaged results for the STS Benchmark \citep{STSBenchmark}, textual entailment, and semantic relatedness \citep{marelli2014semeval} and refer to the average as supervised semantic similarity (SSS). 
We compute the average accuracy for movie review~\citep{pang2005seeing}; customer review~\citep{hu2004mining}; opinion polarity~\citep{wiebe2005annotating}; subjectivity classification~\citep{pang2004sentimental}; Stanford sentiment treebank~\citep{socher2013recursive}; question classification~\citep{li2002learning}; and paraphrase detection~\citep{dolan2004unsupervised}, and refer to it as sentence classification (SC). For the rest of the linguistic probing tasks~\citep{conneau-etal-2018-cram}, we report the average accuracy and report it as ``Probing''.

\subsection{Results}

Table~\ref{table:results} shows the experiment results over all \senteval and \disceval tasks. Different models and training signals have complex effects when performing various downstream tasks. We summarize our findings below: 

\begin{itemizesquish}
    \item On \disceval, \skipthought performs best on RST-DT. \dissent performs strongly for PDTB tasks but it requires discourse markers from PDTB for generating training data. \bert has the highest average by a large margin, but \elmo has competitive performance on multiple tasks.
    
    \item The NL or SPP loss alone has complex effects across tasks in \disceval, but when they are combined, the model achieves the best performance, outperforming our baseline by 0.7\% on average. In particular, it yields 40.5\% accuracy on PDTB-I, outperforming Skip-thought by 0.3\%. This is presumably caused by the differing, yet complementary, effects of these two losses (NL and SPP).
    
    \item The SDT loss generally hurts performance on \disceval, especially on the position-related tasks (SP, BSO). This can be explained by the notion that consecutive sentences in the same section are encouraged to have the same sentence representations when using the SDT loss. However, the SP and BSO tasks involve differentiating neighboring sentences in terms of their position and ordering information. 

    \item On \senteval, SDT is most helpful for the USS tasks, presumably because it provides the most direct information about the topic of each sentence, which is a component of semantic similarity. SDT helps slightly on the SSS tasks. NL gives the biggest improvement in SSS.
    \item In comparing \bert to \elmo and \skipthought to \infersent on \disceval, we can see the benefit of adding information about neighboring sentences. Our proposed training objectives show complementary improvements over NSP, which suggests that they can potentially benefit these pretrained representations.

\end{itemizesquish}

\section{Analysis}

\paragraph{Per-Layer analysis.}
\label{sec:bert-elmo-per-layer-perf}
\begin{figure}
    \centering
    \includegraphics[scale=0.37]{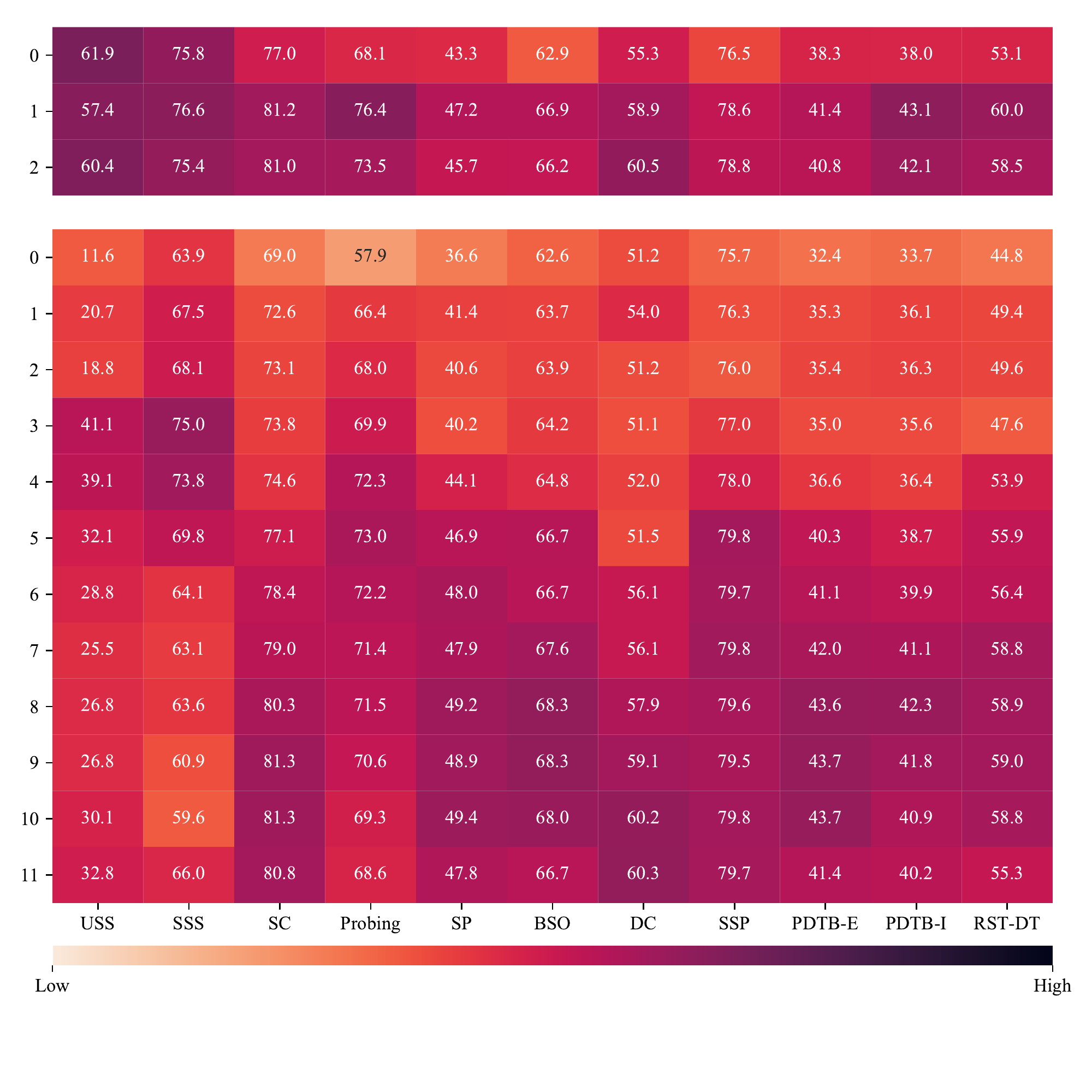}
    \caption{Heatmap for individual hidden layers of \bertbase (lower part) and \elmo (upper part).
    }
    \label{fig:bert_elmo_per_layer_perf}
\end{figure}

\begin{table}[t]
    \small
    \centering
    \begin{tabular}{ccc}
        \toprule
        & \elmo & \bertbase \\
        \midrule
        \senteval & 0.8 & 5.0 \\
        \disceval & 1.3 & 8.9 \\
        \bottomrule
    \end{tabular}
    \caption{Average of the layer number for the best layers in \senteval and \disceval. }
    \label{table:bert_elmo_best_layer}
\end{table}

To investigate the performance of individual hidden layers, we evaluate \elmo and \bert on both \senteval and \disceval using each hidden layer. For \elmo, we use the averaged vector from the targeted layer. For \bertbase, we use the vector from the position of the ``[CLS]'' token. Figure~\ref{fig:bert_elmo_per_layer_perf} shows the heatmap of performance for individual hidden layers. We note that for better visualization, colors in each column are standardized. On \senteval, \bertbase performs better with shallow layers on USS, SSS, and Probing (though not on SC), but on \disceval, the results using \bertbase gradually increase with deeper layers. To evaluate this phenomenon quantitatively, we compute the average of the layer number for the best layers for both \elmo and \bertbase and show it in Table~\ref{table:bert_elmo_best_layer}. 
From the table, we can see that \disceval requires deeper layers to achieve better performance. We assume this is because deeper layers can capture higher-level structure, which aligns with the information needed to solve the discourse tasks.

\paragraph{\disceval architectures.}
In all \disceval tasks except \dcshort, we use no hidden layer in the neural architectures, following the example of SentEval. However, some tasks are unsolvable with this simple architecture. In particular, the \dcshort tasks have low accuracies with all models unless a hidden layer is used. As shown in Table~\ref{table:hidlayer}, when adding a hidden layer of 2000 to this task, the performance on \dcshort improves dramatically. This shows that \dcshort requires more complex comparison and inference among input sentences. Our human evaluation below on \dcshort also shows that human accuracies exceed those of the classifier based on sentence embeddings by a large margin.

\begin{table}[t]
    \small
    \centering
    \begin{tabular}{cc}
        \toprule
        Baseline w/o hidden layer & 52.0 \\
        Baseline w/ hidden layer & 61.0 \\
        \bottomrule
    \end{tabular}
    \caption{Accuracies with baseline encoder on \dc task, with or without a hidden layer in the classifier. \kevin{Did we remove the random baseline? if so, we can cut that last phrase}\mingda{that's weird. removed the phrase.}}
    \label{table:hidlayer}
\end{table}

\begin{table*}[t]
    \small
    \centering
    \begin{tabular}{l|ccc|ccc|cc}
        \toprule
        &\multicolumn{3}{c|}{\spos} & \multicolumn{3}{c|}{\bso} & \multicolumn{2}{c}{\dc} \\
        \midrule
        Human & \multicolumn{3}{c|}{77.3} & \multicolumn{3}{c|}{84.7} & \multicolumn{2}{c}{87.0} \\
        \bertlarge & \multicolumn{3}{c|}{53.8} & \multicolumn{3}{c|}{69.3} & \multicolumn{2}{c}{59.6} \\
        \midrule
                   & Wiki & arXiv & ROC & Wiki & arXiv & ROC & Wiki & Ubuntu \\
        Human      & 84.0 & 76.0  & 94.0 & 64.0 & 72.0  & 96.0  & 98.0  & 74.0 \\
        \bertlarge & 50.7 & 47.3  & 63.4 & 70.4 & 66.8  & 70.8  & 65.1  & 54.2 \\
        \bottomrule
    \end{tabular}
    \caption{Accuracies (\%) for a human annotator and \bertlarge on \spos, \bso, and \dc tasks.}
    \label{table:dc-human}
\end{table*}

\begin{table}[t]
    \small
    \centering
    \begin{tabular}{cl}
        \toprule
        Random & 20 \\
        Baseline w/o context & 43.2 \\
        Baseline w/ context & 47.3 \\
        \bottomrule
    \end{tabular}
    \caption{Accuracies (\%) for baseline encoder on \spos task when using downstream classifier with or without context.}
    \label{table:context}
\end{table}

\paragraph{Human Evaluation.}
We conduct a human evaluation on the \spos, \bso, and \dc datasets. 
A native English speaker was provided with 50 examples per domain for these tasks. While the results in Table~\ref{table:dc-human} show that the overall human accuracies exceed those of the classifier based on \bertlarge by a large margin, we observe that within some specific domains, for example Wiki in BSO, \bertlarge demonstrates very strong performance.

\paragraph{Does context matter in \spos?}
In the \sposshort task, the inputs are the target sentence together with 4 surrounding sentences. We study the effect of removing the surrounding 4 sentences, i.e., only using the target sentence to predict its position from the start of the paragraph.  

Table~\ref{table:context} shows the comparison of the baseline model performance on \spos with or without the surrounding sentences and a random baseline. Since our baseline model is already trained with NSP, it is expected to see improvements over a random baseline. The further improvement from using surrounding sentences demonstrates that the context information is helpful in determining the sentence position.

\section{Conclusion}

We proposed \disceval, a test suite of tasks to evaluate discourse-related knowledge encoded in pretrained sentence representations.
We also proposed a variety of training objectives to strengthen encoders' ability to incorporate discourse information. 
We benchmarked several pretrained sentence encoders and demonstrated the effects of the proposed training objectives on different tasks.
While our learning criteria showed benefit on certain classes of tasks, our hope is that the \disceval evaluation suite can inspire additional research in capturing broad discourse context in fixed-dimensional sentence embeddings.

\section*{Acknowledgments}
We thank Jonathan Kummerfeld for helpful discussions about the IRC Disentanglement dataset, Davis Yoshida for discussions about \bert, and the anonymous reviewers for their feedback that improved this paper.  
This research was supported in part by a Bloomberg data science research grant to K.~Gimpel.

\bibliography{acl2019}
\bibliographystyle{acl_natbib}

\end{document}

\maketitle

\appendix

\begin{table}[t]
\small
\centering
\begin{tabular}{c}
\toprule 
RST-DT \\
\midrule
Attribution\\
Background\\
Cause\\
Comparison\\
Condition\\
Contrast\\
Elaboration\\
Enablement\\
Evaluation\\
Explanation\\
Joint\\
Manner-Means\\
Same-unit\\
Summary\\
Temporal\\
Textual-organization\\
Topic-Change\\
Topic-Comment\\
\bottomrule
\end{tabular}
\vspace{-0.05cm}
\caption{\label{table:rst-dt-labels} 18 coarse-grained relations in RST-DT}
\vspace{-0.1cm}
\end{table}

\begin{table}[t]
\small
\centering
\begin{tabular}{l|l}
\toprule 
PDTB-E & PDTB-I \\
\midrule
Comparison.Concession & Comparison.Concession\\
Comparison.Contrast & Comparison.Contrast\\
Contingency.Cause & Contingency.Cause\\
Contingency.Condition & Contingency.Prag cause\\
Contingency.Prag condition & Expansion.Alternative\\
Expansion.Alternative & Expansion.Conjunction\\
Expansion.Conjunction&Expansion.Instantiation\\
Expansion.Instantiation&Expansion.List\\
Expansion.List&Expansion.Restatement\\
Expansion.Restatement&Temporal.Asynchronous\\
Temporal.Asynchronous&Temporal.Synchrony\\
Temporal.Synchrony&\\
\bottomrule
\end{tabular}
\vspace{-0.05cm}
\caption{\label{table:pdtb-labels} The PDTB relation categories}
\vspace{-0.1cm}
\end{table}

\begin{figure}[t]
    \centering
    \includegraphics[scale=0.24]{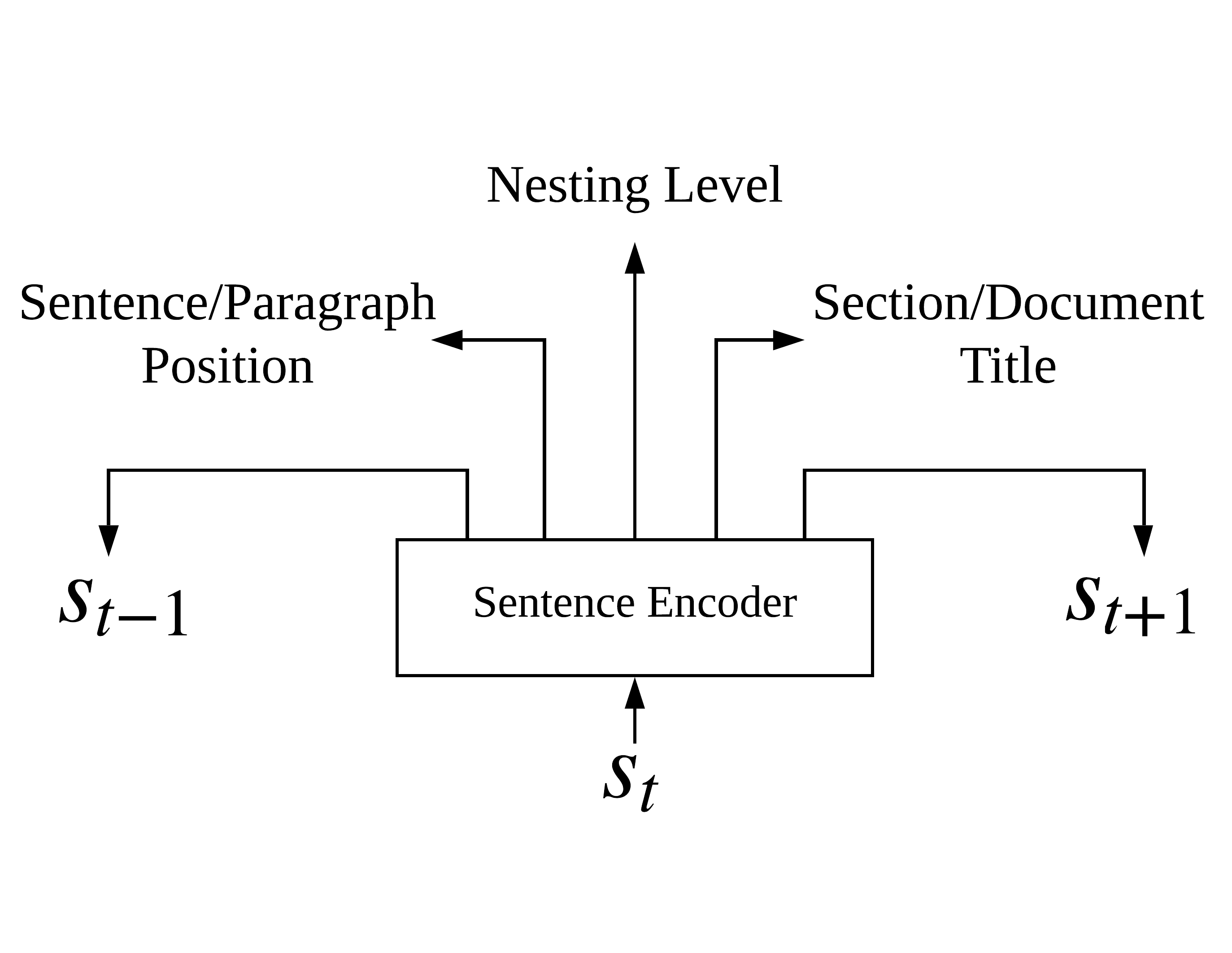}
    \vspace{-0.7cm}
    \caption{Schematic showing multitask training for our sentence embedding model.}
    \label{fig:model}
\end{figure}

\section{Hyperparameters}

Our models use 1200 dimensional BiGRUs, resulting in 2400 dimensional sentence representations. The feedforward neural networks used in the decoders are parameterized using two hidden layers and use ReLU activation functions. We intialize our models with 300 dimensional GloVe embeddings~\cite{glove}. We use Adam~\cite{kingma2014adam} as optimizer and train our models for one epoch on Wikipedia without employing early stopping.

\bibliography{acl2019}
\bibliographystyle{acl_natbib}